\documentclass[11pt]{article}
\usepackage{a4}       % Estils de ACL
\usepackage{acl}      % Estils de ACL
\usepackage{amssymb}  % simbols de AMS
\usepackage{epsf}     % figures en PostScript
\usepackage{array}    % per les opcions del tabular
\usepackage{macros}   % les meves macros

\title{A Comparison between Supervised Learning Algorithms 
       for Word Sense Disambiguation} 

\author{
  {\large\bf Gerard Escudero},{\large\bf\ \ Llu\'{\i}s M\`arquez}, 
  {\large {\rm and} {\bf German Rigau}\thanks{This research has been 
   partially funded by the Spanish Research Department (CICYT's
   project TIC98--0423--C06), by the EU Commission (NAMIC 
   IST-1999-12392), and by the Catalan Research Department 
   (CIRIT's consolidated research group 1999SGR-150 and CIRIT's 
   grant 1999FI 00773).}}\\
  TALP Research Center. LSI Department. 
  Universitat Polit\`ecnica de Catalunya (UPC)\\
  Jordi Girona Salgado 1--3. E-08034 Barcelona. Catalonia\\
  {\tt \{escudero,lluism,g.rigau\}@lsi.upc.es}}

\begin{document}
%----------------------------------------- T i t o l
\maketitle
\pagestyle{empty}

%----------------------------------------- R e s u m
\begin{abstract}
  This paper describes a set of comparative experiments, including
  cross--corpus evaluation, between five alternative algorithms for
  supervised Word Sense Disambiguation (\aWSD), namely Naive Bayes,
  Exemplar-based learning, \aSNe, Decision Lists, and Boosting. Two
  main conclusions can be drawn: 1) The \aLBe\ algorithm outperforms
  the other four state-of-the-art algorithms in terms of accuracy and
  ability to tune to new domains; 2) The domain dependence of \aWSD\
  systems seems very strong and suggests that some kind of adaptation
  or tuning is required for cross--corpus application. 
%%%%%%%%%%%%%%%%%
%  Furthermore, the rules produced by \aLBe\ 
%  helps understanding the differences between corpora and might
%  help improving data quality.
%  \medskip\\ 
%  {\bf Keywords:} Portability and Tuning of \aNLP\ systems, Word Sense
%  Disambiguation, Supervised Machine Learning.\\ 
\end{abstract}

%----------------------------------------- I n t r o d u c c i o
\section{Introduction}
Word Sense Disambiguation (\aWSD) is the problem of assigning the
appropriate meaning (or sense) to a given word in a text or discourse. 
Resolving the ambiguity of words is a central problem for large 
scale language understanding applications and their associate 
tasks~\cite{ide98}. 
Besides, \aWSD\ is one of the most important open problems in 
\aNLP. Despite the wide range of approaches 
investigated~\cite{kilgarriff00} and the large effort devoted 
to tackle this problem, to date, no large-scale broad-coverage
and highly accurate \aWSD\ system has been built.

One of the most successful current lines of research is the
corpus-based approach in which statistical or Machine Learning (\aML)
algorithms have been applied to learn statistical models or
classifiers from corpora in order to perform \aWSD. Generally,
supervised approaches (those that learn from previously semantically
annotated corpus) have obtained better results than unsupervised
methods on small sets of selected ambiguous words, or artificial
pseudo-words. Many standard \aML\ algorithms for supervised learning
have been applied, such as: Decision Lists~\cite{yarowsky94,agirre00},
Neural Networks~\cite{towell98}, Bayesian learning~\cite{bruce99}, 
Exemplar-Based learning~\cite{ng97a}, and
Boosting~\cite{escudero00a}, etc. Further, in~\cite{mooney96} some of
the previous methods are compared jointly with Decision Trees
and Rule Induction algorithms, on a very restricted domain.

Although some comparative studies between alternative algorithms 
have been reported~\cite{mooney96,ng97a,escudero00a,escudero00b},
none of them addresses the issue of the portability of supervised \aML\
algorithms for \aWSD, i.e. to test whether the accuracy of a system 
trained on a certain corpus can be extrapolated to other corpora or not.
We think that the study of the domain dependence of \aWSD\ ---in the
style of other studies devoted to parsing
\cite{sekine97,ratnaparkhi99}--- is needed to assess the validity of
the supervised approach, and to determine to which extent a
pre--process of tuning is necessary to make real \aWSD\ systems
portable.  In this direction, this work compares five different \aML\ 
algorithms and explores their portability and tuning ability by
training and testing them on different corpora.

%----------------------------------------- L e a r n i n g   A l g or i t h m s
\section{Learning Algorithms Tested}
\label{s-MLmethods}

%----------------------------------- Naive Bayes
\paragraph{Naive-Bayes (\aNB).}
Naive Bayes is intended as a simple representative of statistical
learning methods. It has been used in its most classical
setting~\cite{duda73}. That is, assuming independence of features, it
classifies a new example by assigning the class that maximizes the
conditional probability of the class given the observed sequence of
features of that example.

Model probabilities are estimated during training process using
relative frequencies. To avoid the effect of zero counts when
estimating probabilities, a very simple smoothing technique has 
been used, which was proposed in~\cite{ng97a}. 
%%%%
%In our implementation \cite{escudero00b}, when classifying new
%examples only the information of words appearing in the example
%(positive information) is taken into account. In that paper, these
%variants are called {\it positive} Naive Bayes (\aPNB).
%%%%

Despite its simplicity, Naive Bayes is claimed to obtain
state--of--the--art accuracy on supervised \aWSD\ in many
papers~\cite{mooney96,ng97a,leacock98}.\medskip

%----------------------------------- Exemplar Based
\noindent{\bf Exemplar-based Classifier (\aEB).}\ \ In exemplar,
instance, or memory--based learning \cite{aha91} no generalization of
training examples is performed. Instead, the examples are stored in
memory and the classification of new examples is based on the classes
of the most similar stored examples.  In our implementation,
%\cite{escudero00b}, 
all examples are kept in memory and the classification of a new
example is based on a $k$--NN (Nearest--Neighbours) algorithm using
Hamming distance to measure closeness. For $k$'s greater than 1, the
resulting sense is the weighted majority sense of the $k$ nearest
neighbours ---where each example votes its sense with a strength
proportional to its closeness to the test example.

%%%%
%The topical information is codified as a single set--valued
%attribute (containing all words appearing in the sentence) and the
%calculation of closeness is modified so as to handle this type of
%attribute. In that paper, these variants are called {\it positive}
%Exemplar--based (\aPEB).
%%%%

Exemplar--based learning is said to be the best option for
\aWSD~\cite{ng97a}.  Other authors~\cite{daelemans99} point out that
exemplar--based methods tend to be superior in language learning
problems because they do not forget exceptions.

%----------------------------------- Snow
\paragraph{SNoW: A Winnow--based Classifier.}
\aSNe\ stands for Sparse Network Of Winnows, and it is intended as a
representative of on--line learning algorithms.  In
the \aSNe\ architecture there is a Winnow \cite{littlestone88} node
for each class, which learns to separate that class from all the rest.
In this paper, our approach to \aWSD\ using \aSNe\ follows that
of~\cite{escudero00c}.

\aSNe\ is proven to perform very well in high dimensional domains,
where both, the training examples and the target function reside very 
sparsely in the feature space \cite{roth98b}, e.g: text categorization,
context--sensitive spelling correction, \aWSD, etc.

%----------------------------------- Decision Lists
\paragraph{Decision Lists (\aDL).} In this setting, \aDLe\ are 
ordered lists of features extracted from the training examples and
weighted by a log--likelihood measure \cite{yarowsky94}. The
aproximation described in \cite{agirre00} has been fully used (using
also their pruning and smoothing techniques).

\aDLe\ were one of the most succesful systems on the 1st edition of
the Senseval competition \cite{kilgarriff00}.

%----------------------------------- Boosting
\paragraph{LazyBoosting (\aLB).} The main idea of boosting algorithms 
is to combine many simple and moderately accurate hypotheses (called
weak classifiers) into a single, highly accurate classifier. The weak
classifiers are trained sequentially and, conceptually, each of them
is trained on the examples which were most difficult to classify by
the preceding weak classifiers.

\aLBe~\cite{escudero00a}, is a simple modification of the \aABMH\ 
algorithm \cite{schapire00a}, which consists of reducing the feature
space that is explored when learning each weak classifier. More
specifically, a small proportion of attributes are randomly selected
and the best weak rule is selected only among them.  This modification
significantly increases the efficiency of the learning process with no
loss in accuracy.

%----------------------------------------- S e t t i n g
\section{Setting}
\label{s-setting}

The set of comparative experiments has been carried out on a subset of
21 words of the \aDSO\ corpus, which is a semantically annotated
English corpus collected by Ng and colleagues~\cite{ng96}, and
available from the Linguistic Data Consortium (LDC)\footnote{
%LDC address: 
  {\tt http://www.ldc.upenn.edu/}}.  Each word is treated as
a different classification problem. They are 13 nouns ({\sf age, art,
  body, car, child, cost, head, interest, line, point, state, thing,
  work}) and 8 verbs ({\sf become, fall, grow, lose, set, speak,
  strike, tell}). The average number of senses per word is close to 10
and the number of training examples is close to 1,000.

The \aDSO\ corpus contains sentences from two different corpora,
namely Wall Street Journal (\aWSJ) and Brown Corpus (\aBC).
Therefore, it is easy to perform experiments about the portability of
alternative systems by training them on the \aWSJ\ part (\A\ part,
hereinafter) and testing them on the \aBC\ part (\B\ part,
hereinafter), or vice-versa.\medskip

%----------------------------------------- 
Two kinds of information are used to train classifiers: local 
and topical context. The former consists of the words and 
part-of-speech tags appearing in a window of $\pm$ 3 items 
around the target word, and collocations of up to three 
consecutive words in the same window. The latter consists 
of the unordered set of content words appearing in the whole 
sentence.

%----------------------------------------- E x p e r i m e n t s
\section{Experiments}
\label{s-experiments}

%----------------------------------------- 
\subsection{Comparing the five approaches}

The five algorithms, jointly with a naive Most-Frequent-sense
Classifier (\aMFC), have been tested on 7 different combinations of
training--test sets\footnote{The combinations of training--test sets
  are called: \ambsamb, \ambsa, \ambsb, \asa, \bsb, \asb, and \bsa,
  respectively.  In this notation, the training set is placed at the
  left hand side of symbol ``--'', while the test set is at the right
  hand side. For instance, \asb\ means that the training set is corpus
  \A\ and the test set is corpus \B. The symbol ``$+$'' stands for set
  union.}.
Accuracy figures, averaged over the
21 words, are reported in table~\ref{t-results1st}.  The comparison
leads to the following conclusions:\smallskip

\begin{table*}[htb]
\begin{tabular}{l|c|c|c||c|c||c|c|}\cline{2-8}
       & \multicolumn{7}{c|}{{\sf Accuracy {\small (\%)}}}\\\cline{2-8}
       & \sambsamb & \sambsa & \sambsb & \sasa & \sbsb & \sasb & \sbsa \\\hline
%\multicolumn{1}{|l|}{\aMFC}          & 46.55 & 53.90 & 39.21 & 55.94 & 45.52 & 36.40 & 38.71 \\\hline
%\multicolumn{1}{|l|}{Naive Bayes}    & 61.55 & 67.25 & 55.85 & 65.86 & 56.80 & 41.38 & 47.66 \\\hline
%\multicolumn{1}{|l|}{Exemplar--based}& 63.01 & 69.08 & 56.97 & 68.98 & 57.36 & 45.32 & 51.13 \\\hline
%\multicolumn{1}{|l|}{\aDLe}          & 61.58 & 67.64 & 55.53 & 67.57 & 56.56 & 43.01 & 48.83 \\\hline
%\multicolumn{1}{|l|}{\aSNe}          & 60.92 & 65.57 & 56.28 & 67.12 & 56.13 & 44.07 & 49.76 \\\hline
%\multicolumn{1}{|l|}{\aLBe}          & {\bf 66.32} & {\bf 71.79} & {\bf 60.85} & {\bf 71.26} & {\bf 58.96} & {\bf 47.10} & {\bf 51.99}$^*$ \\\hline

\multicolumn{1}{|l|}{\aMFC}          & 46.55{\scs$\pm$0.71} & 53.90{\scs$\pm$2.01} & 39.21{\scs$\pm$1.90} & 55.94{\scs$\pm$1.10} & 45.52{\scs$\pm$1.27} & 36.40 & 38.71 \\\hline
\multicolumn{1}{|l|}{Naive Bayes}    & 61.55{\scs$\pm$1.04} & 67.25{\scs$\pm$1.07} & 55.85{\scs$\pm$1.81} & 65.86{\scs$\pm$1.11} & 56.80{\scs$\pm$1.12} & 41.38 & 47.66 \\\hline
\multicolumn{1}{|l|}{Exemplar--based}& 63.01{\scs$\pm$0.93} & 69.08{\scs$\pm$1.66} & 56.97{\scs$\pm$1.22} & 68.98{\scs$\pm$1.06} & 57.36{\scs$\pm$1.68} & 45.32 & 51.13 \\\hline
\multicolumn{1}{|l|}{\aDLe}          & 61.58{\scs$\pm$0.98} & 67.64{\scs$\pm$0.94} & 55.53{\scs$\pm$1.85} & 67.57{\scs$\pm$1.44} & 56.56{\scs$\pm$1.59} & 43.01 & 48.83 \\\hline
\multicolumn{1}{|l|}{\aSNe}          & 60.92{\scs$\pm$1.09} & 65.57{\scs$\pm$1.33} & 56.28{\scs$\pm$1.10} & 67.12{\scs$\pm$1.16} & 56.13{\scs$\pm$1.23} & 44.07 & 49.76 \\\hline
\multicolumn{1}{|l|}{\aLBe}          & {\bf 66.32}{\scs$\pm$1.34} & {\bf 71.79}{\scs$\pm$1.51} & {\bf 60.85}{\scs$\pm$1.81} & {\bf 71.26}{\scs$\pm$1.15} & {\bf 58.96}{\scs$\pm$1.86} & {\bf 47.10} & {\bf 51.99}$^*$ \\\hline
\end{tabular}
\caption{Accuracy results ($\pm$ standard deviation) of the methods on all training--test combinations}
\label{t-results1st}
\end{table*}

\aLBe\ outperforms the other three methods in all tests. The
difference is statistically significant in all cases except when
comparing \aLBe\ to the \aEBe\ approach in the case marked with an
asterisk\footnote{Statistical tests of significance applied: McNemar's
  test and 10-fold cross-validation paired Student's $t$-test at a
  confidence value of 95\%~\cite{dietterich98}.}.\smallskip

Extremely poor results are observed when testing the portability of
the systems. Restricting to \aLBe\ results, we observe that the
accuracy obtained in \asb\ is 47.1\% while the accuracy in \bsb\ 
(which can be considered an upper bound for \aLBe\ in \B\ corpus) is
59.0\%, that is, a drop of 12 points. Furthermore, 47.1\% is only
slightly better than the most frequent sense in corpus \B,
45.5\%.\smallskip

Apart from accuracy figures, the observation of the predictions made
by the five methods on the test sets provides interesting information
about the comparison of the algorithms. Table~\ref{t-kappa} shows the
agreement rates and the Kappa ($\kappa$) statistics\footnote{The Kappa
  statistic $k$ \cite{cohen60} is a better measure of inter--annotator
  agreement which reduces the effect of chance agreement. It has been
  used for measuring inter--annotator agreement during the
  construction of some semantic annotated corpora
  \cite{veronis98,ng99}.} between all pairs of methods in the
\ambsamb\ case. `\aDSO' stands for the annotation of \aDSO\ corpus,
which is taken as the correct. Therefore the agreement rate with
\aDSO\ contains the accuracy results previously reported. Some
interesting conclusions can be drawn from those tables:

\paragraph{1.} \aNB\ obtains the most similar results with regard to \aMFC\ in
agreement rate and Kappa values in all tables. The agreement ratio is
76\%, that is, more than 3 out of 4 times it predicts the most
frequent sense.

\paragraph{2.} \aLB\ obtains the most similar results with regard to \aDSO\ 
(accuracy) in agreement rate and Kappa values, and it has the less
similar Kappa and agreement values with regard to \aMFC\ 
%(furthermore, \aLB\ provides the most dissimilar annotation with respect to the rest
%of algorithms)
. This indicates that \aLB\ is the method that better
learns the behaviour of the \aDSO\ examples.

\paragraph{3.} The Kappa values are very low. But, as it is suggested in
\cite{veronis98}, evaluation measures, such as precision and recall,
should be computed relative to the agreement between the human
annotators of the corpus and not to a theoretical 100\%. It seems
pointless to expect more agreement between the system and the
reference corpus than between the annotators themselves. Contrary to
the intuition that the agreement between human annotators should be
very high in the \aWSD\ task, some papers report surprisingly low
figures. For instance, \cite{ng99} reports an accuracy rate of 56.7\%
and a Kappa value of 0.317 when comparing the annotation of a subset
of the \aDSO\ corpus performed by two independent research
groups\footnote{A Kappa value of 1 indicates perfect agreement, while
  0.8 is considered as indicating good agreement \cite{carletta96}.}.
Similarly, \cite{veronis98} reports values of Kappa near to zero when
annotating some special words for the ROMANSEVAL
corpus\footnote{
%ROMANSEVAL is, like SENSEVAL for English, a specific
%  competition between \aWSD\ systems for Romance languages.  See
  http://www.lpl.univ-aix.fr/projects/romanseval 
%and http://www.itri.bton.ac.uk/events/senseval.
}. From this point of
view, the Kappa values of 0.44 achieved by \aLB\ in \ambsamb\ could be
considered excellent results.  Unfortunately, the subset of the \aDSO\ 
corpus and that used in this report are not the same and, therefore, a
direct comparison is not possible.

\begin{table}[htb]\centering
{\footnotesize\begin{tabular}{l|c|c|c|c|c|c|c|}\cline{2-8}
& \multicolumn{7}{c|}{\sambsamb}\\\cline{2-8}
& \aGS & \aMFC & \aNB & \aEB & \aSN & \aDL & \aLB\\\hline
\multicolumn{1}{|c|}{\aGS}  & ---  & 46.6 & 61.6 & 63.0 & 60.9 & 61.6 & 66.3 \\\hline
\multicolumn{1}{|c|}{\aMFC} &-0.19 & ---  & 73.9 & 60.0 & 55.9 & 64.9 & 54.9 \\\hline
\multicolumn{1}{|c|}{\aNB}  & 0.24 &-0.09 & ---  & 76.3 & 74.5 & 76.8 & 71.4 \\\hline
\multicolumn{1}{|c|}{\aEB}  & 0.36 &-0.15 & 0.44 & ---  & 69.6 & 70.7 & 72.5 \\\hline
\multicolumn{1}{|c|}{\aSN}  & 0.36 &-0.17 & 0.44 & 0.44 & ---  & 67.5 & 69.0 \\\hline
\multicolumn{1}{|c|}{\aDL}  & 0.32 &-0.13 & 0.40 & 0.41 & 0.38 & ---  & 69.9 \\\hline
\multicolumn{1}{|c|}{\aLB}  & 0.44 &-0.17 & 0.37 & 0.50 & 0.46 & 0.42 & ---  \\\hline
\end{tabular}}
\caption{Kappa ($\kappa$) statistic (below diagonal) and agreement
  rate (above diagonal) between all methods in \ambsamb\ experiments}
\label{t-kappa}
\end{table}

%----------------------------------------- 
\subsection{About the tuning to new domains}
This experiment explores the effect of a simple tuning process
consisting of adding to the original training set a relatively small
sample of manually sense tagged examples of the new domain. The size
of this supervised portion varies from 10\% to 50\% of the available
corpus in steps of 10\% (the remaining 50\% is kept for testing).
Results indicate that: \aLBe\ is again superior to their competitors. 

\begin{figure*}[htbp]\centering
\begin{tabular}{ccc}
(a) \aNBe & (b) \aEBe & (c) \aSNe\\
\hspace*{-1cm}\epsfxsize=6cm\epsfbox{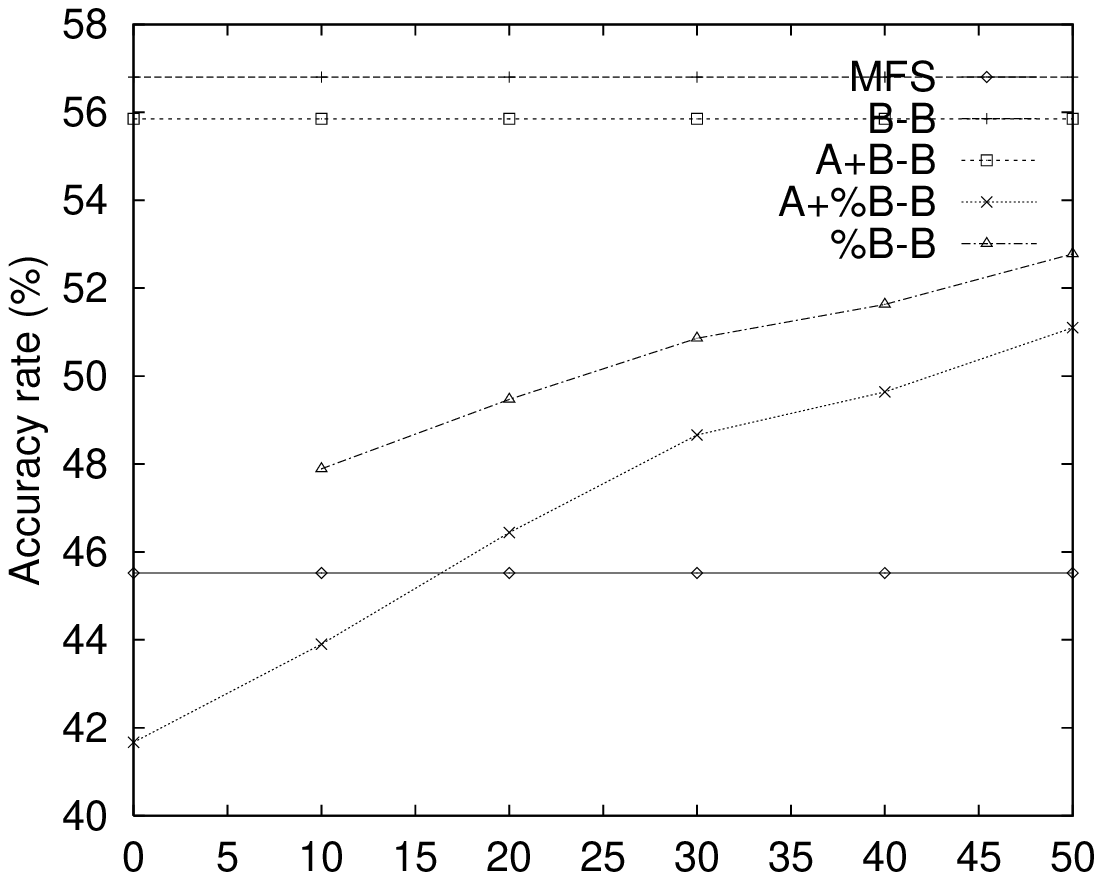}&
\hspace*{-8mm}\epsfxsize=6cm\epsfbox{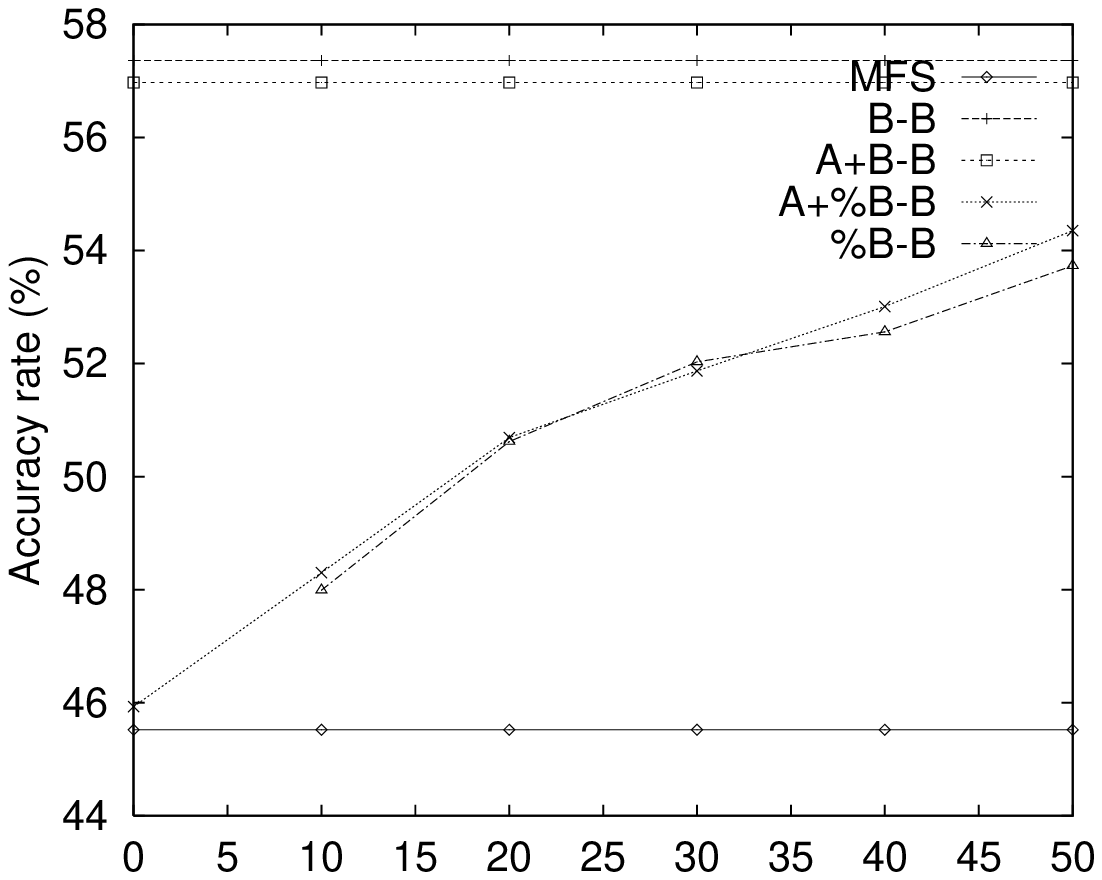} &
\hspace*{-8mm}\epsfxsize=6cm\epsfbox{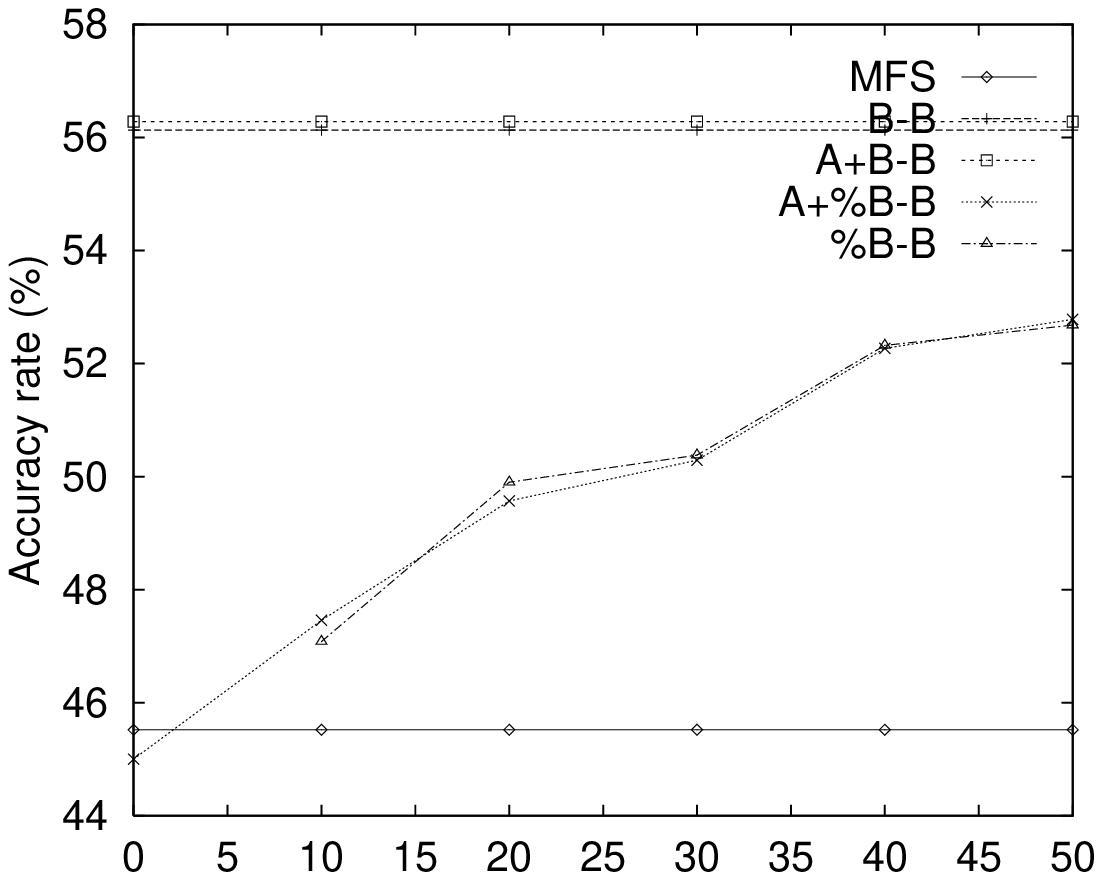}
\smallskip\\
\end{tabular}
\begin{tabular}{cc}
(d) \aDLe & (e) \aLBe\\
\hspace*{-1cm}\epsfxsize=6cm\epsfbox{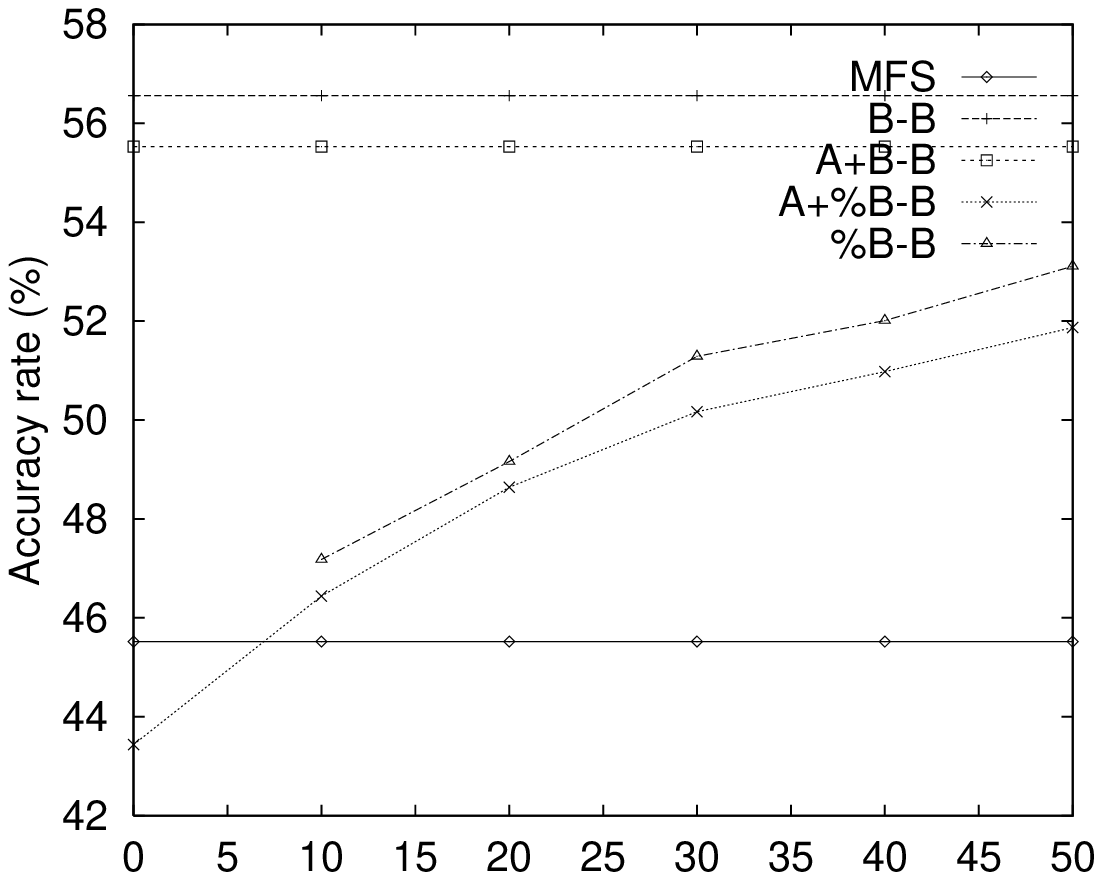} &
\hspace*{-5mm}\epsfxsize=6cm\epsfbox{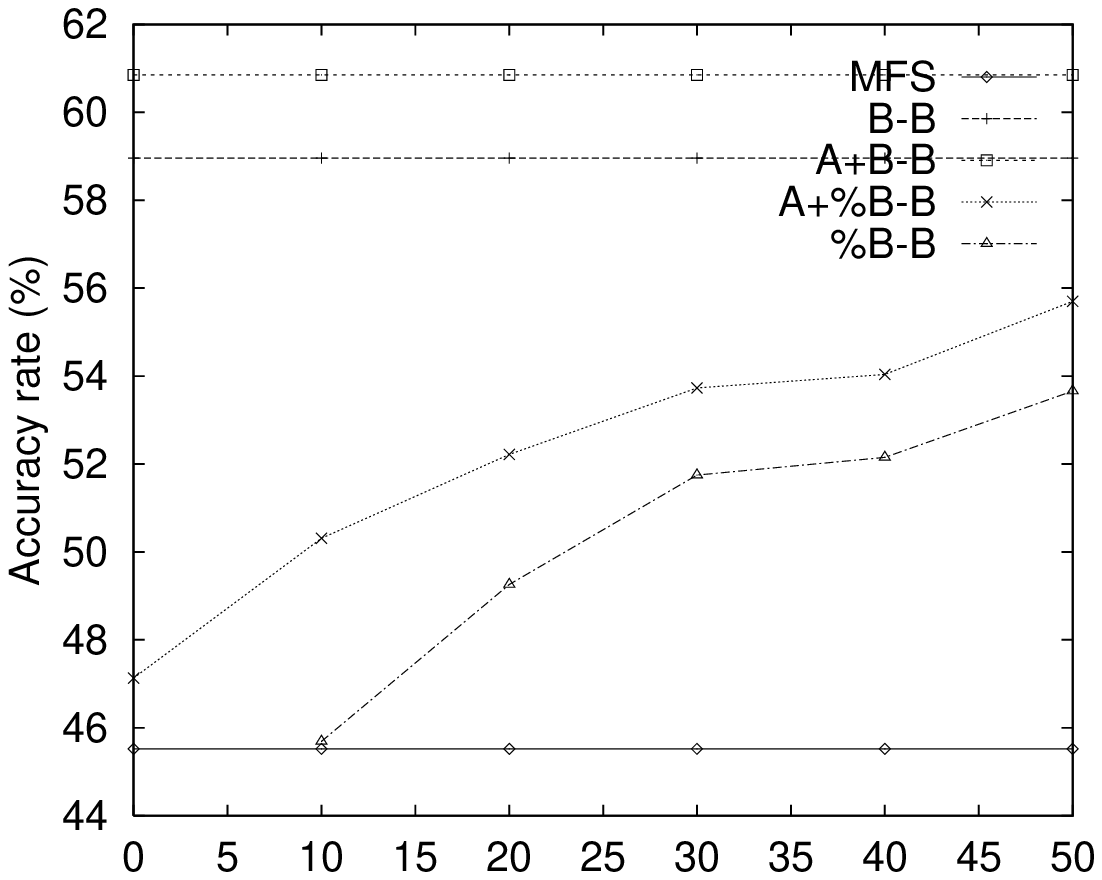}
\end{tabular}
\caption{Results of the tuning experiment}
\label{f-results2nd}
\end{figure*}

Summarizing, the results obtained show that for \aNBe, \aEBe, \aSNe\ 
and \aDLe\ methods it is not worth keeping the original training
examples. Instead, a better (but disappointing) strategy would be
simply using the tuning corpus.  However, this is not the situation of
\aLBe, for which a moderate (but consistent) improvement of accuracy
is observed when retaining the original training set.\smallskip

We observed that part of the poor results obtained is explained by: 1)
Corpus \A\ and \B\ have a very different distribution of senses, and,
therefore, different a--priori biases; Furthermore, 2) Examples of
corpus \A\ and \B\ contain different information, and, therefore, the
learning algorithms acquire different (and non interchangeable)
classification cues from both corpora. The study of the rules acquired
by \aLBe\ from \aWSJ\ and \aBC\ helped understanding the differences
between corpora.  On the one hand, the type of features used in the
rules were significantly different between corpora, and, additionally,
there were very few rules that apply to both sets; On the other hand,
the sign of the prediction of many of these common rules was somewhat
contradictory between corpora \cite{escudero00c}.

%%%
%Both previous experiments give evidence towards the necessity of
%defining effective strategies for: 1) constructing representative and
%useful training corpora; and 2) tuning supervised algorithms.
%%%

%----------------------------------------- 
\subsection{About the training data quality}
The observation of the rules acquired by \aLBe\ also could help
improving data quality. It is known that mislabelled examples
resulting from annotation errors tend to be hard examples to classify
correctly, and, therefore, tend to have large weights in the final
distribution. This observation allows both to identify the noisy
examples and use \aLBe\ as a way to improve the training corpus.

A preliminary experiment has been carried out in this direction by
studying the rules acquired by \aLBe\ from the training examples of
word {\it state}.  The manually revision of the 50 highest scored
rules allowed us to identify a high number of noisy training examples
--there were 11 of 50 tagging errors--, and, additionally, 17 examples
of 50 not coherently tagged, probably due to the too fine grained or
not so clear distinctions between the senses involved in these
examples. Thus, there were 28 of 50 examples with some problem, that
is more than 1 of each two cases have a problem.

%----------------------------------------- C o n c l u s i o n s  &
%                                          F u r t h e r    W o r k
\section{Conclusions and Future Work}
\label{s-conclusions}
This work reports a comparative study of five \aML\ approaches to
\aWSD, and focuses on studying their portability.  The main
conclusions are:\smallskip

\aLBe\ algorithm outperforms the other four state-of-the-art
supervised \aML\ methods in all domains tested. Furthermore, this
algorithm shows better properties when is tuned to new domains.
\smallskip

Portability is a very important issue that has been paid little
attention up to the present. In this paper we show that a process of
tuning to the domain of application is required to assure the
portability of \aWSD\ systems (at least if the learning--testing
corpora differ so as \aBC\ and \aWSJ\ do).  This evidence questions
the idea of ``robust broad-coverage \aWSD'' introduced
by~\cite{ng97b}, in which a supervised system trained on a large
enough corpora (say a thousand examples per word) should provide
fairly accurate disambiguation on any corpora. To determine the
viability of the supervised approach to \aWSD\ we belief that a
serious effort should be devoted to study the problem of obtaining
representative enough training corpora at a reasonable cost.

Further work is planned to be done in the following directions:

\paragraph{1.} Since most of the knowledge learned from a domain is not 
useful when changing to a new domain, further investigation is needed
on tuning strategies, specially on those using non-supervised
algorithms.

\paragraph{2.} It has been noted that mislabelled examples resulting from
annotation errors tend to be hard examples to classify correctly, and,
therefore, tend to have large weights in the final distribution. It
could provide the methodologies to automatic verify the semantic
annotation of corpora and the grouping of senses.

\paragraph{3.} Moreover, the inspection of the rules learned by
\aLBe\ could provide evidence about similar behaviours of a--priori
different senses. This type of knowledge could be useful to perform
clustering of too fine-grained or artificial senses.

%%----------------------------------------- A g r a i m e n t s
%\section*{Acknowledgments}
%{\small This research has been partially funded by the Spanish
%  Research Department (CICYT's project TIC98--0423--C06), by the EU
%  Comission (NAMIC IST-1999-12392), and by the Catalan Research
%  Department (CIRIT's consolidated research group 1999SGR-150 
%  and CIRIT's grant 1999FI 00773).}

%----------------------------------------- B i b l i o g r a f i a
\bibliographystyle{acl}
%{\small\bibliography{/usr/usuaris/ia/lluism/publicacions/bibliografia/wsd+ml}}
{\small\bibliography{/usr/usuaris/ia/escudero/papers/bibliografia/bibliografia}}

\end{document}